\def\year{2021}\relax
\documentclass[letterpaper]{article} 
\usepackage{aaai21}  
\usepackage{xcolor}
\usepackage{times}  
\usepackage{helvet} 
\usepackage{courier}  
\usepackage[hyphens]{url}  
\usepackage{graphicx} 
\usepackage{natbib}  
\usepackage{caption} 
\usepackage{amsmath}
\usepackage{subcaption}
\usepackage{multirow}
\usepackage{longtable}
\usepackage{bm}
\usepackage{array}

\title{Plug-and-Blend: A Framework for Plug-and-play Controllable Story Generation with Sketches}
\author {
    Zhiyu Lin,\textsuperscript{\rm 1}
    Mark O. Riedl \textsuperscript{\rm 1}
    \\
}
\affiliations {
    \textsuperscript{\rm 1} Georgia Institute of Technology \\
    zhiyulin@gatech.edu, riedl@cc.gatech.edu
}

\begin{document}

\maketitle

\begin{abstract}
Large pre-trained neural language models (LM) have very powerful text generation capabilities. 
However, in practice, they are hard to control for creative purposes.
We describe a Plug-and-Play controllable language generation framework, Plug-and-Blend, that 
allows a human user to input
multiple control codes (topics).
In the context of automated story generation, this allows a human user loose or fine-grained control of the topics and transitions between them that will appear in the generated story, and can even allow for overlapping, blended topics.
Automated evaluations show our framework, working with different generative LMs,
controls the generation towards given continuous-weighted control codes while keeping the generated sentences fluent, demonstrating strong blending capability.
A human participant evaluation shows that the generated stories are observably transitioning between two topics.

\end{abstract}

\section{Introduction}


\begin{figure*}
    \centering
    \includegraphics[width=\textwidth]{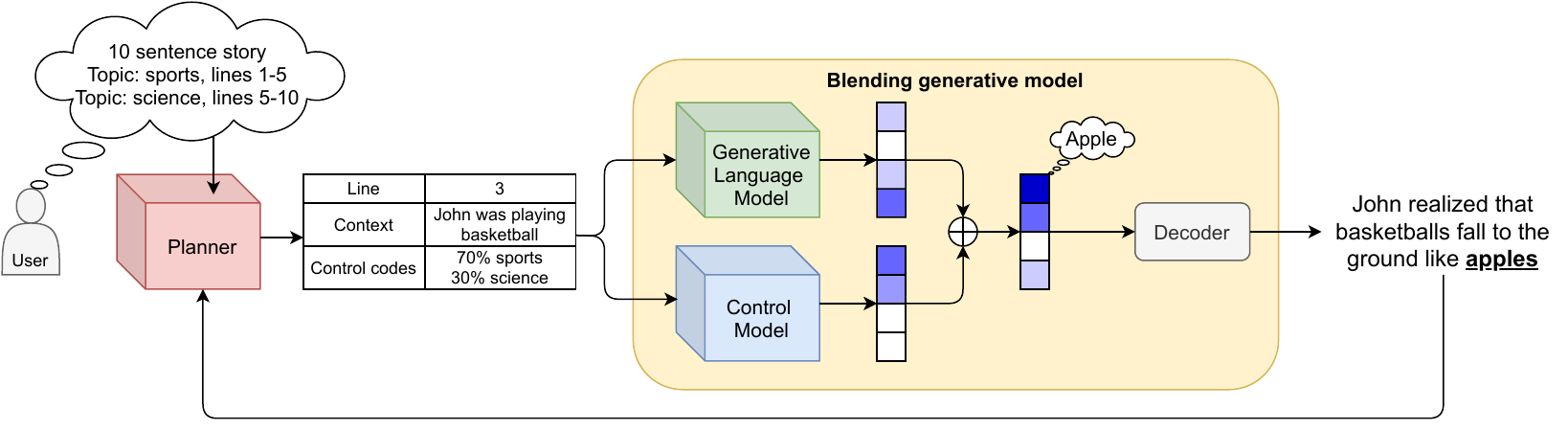}
    \caption{Illustration of overall architecture of our framework}
    \label{fig:overall}
\end{figure*}

Recent advancement in very large pre-trained neural language models (PLM)
have enabled a new generation of Procedural Content Generation (PCG) applications that make use of the generation capability they provide.
However, these PLMs are also difficult to {\bf control} beyond providing a prompt for a generated continuation.
This makes these PLMs ill-suited for {\em co-creative} PCG tasks wherein a game creator works with a language model iteratively to produce novel content, such as stories.
Co-creative tasks require an ability to not only prompt the model but to guide the generation with high-level features, for example, style, context, or topic constraints, that is more intuitive to content creators.


{\em Conditional generation} is a family of generation methods that attempt to provide control "knobs” for human creators.
These techniques either train or fine-tune models in whole
~\citep{hu_toward_2017,fang_transformer-based_2021,keskar_ctrl_2019},
or by sideloading additional discriminators along with a pre-trained model, without changing base model parameters holistically \citep{dathathri_plug_2020,madotto_plug-and-play_2020,duan_pre-train_2020,mai_plug_2020,liu_--fly_2021}.

We seek ``plug-and-play'' approaches to controllable generation wherein new language models can be slotted into existing generative systems;
new language models are being developed and it becomes intractable to update and retrain controlled generation architectures.
Plug-and-play techniques such as \citep{krause_gedi_2020,pascual_directed_2020} aim to only intervene with the outputs---a vector of logits---of a generative language model.
This becomes 
important as the latest iteration of very large PLMs such as GPT-3 
~\cite{brown_language_2020} 
restrict access to the hidden states and layer weights of models.
As language models improve, they can be easily incorporated into existing, controllable generation frameworks.



We present {\em Plug-and-Blend},
\footnote{Code available at \url{https://github.com/xxbidiao/plug-and-blend}}
an efficient plug-and-play generative framework for controllable text generation that
(a)~works with the logit outputs of any language model;
(b)~facilitates fine control of generated sentences by allowing continuous steering towards specific control codes; and
(c)~allows multiple control codes representing style and topic constraints to be provided in overlapping contexts.
These control codes can be blended together to generate content that meets multiple style or topic constraints.
We describe that these key capabilities empower latent space walking in the hyperspace of generated sentences, and show a simple content planning technique that utilizes this feature to generate stories regarding user intentions in a co-authoring. 

We present our approach in the context of PCG and automated story generation, and show its capability of both fine- and high-level blending demonstrated in automated and human subject evaluations. 

    


    

\section{Related Work}

\subsubsection{Plug-and-Play Conditional Generation}
\label{subsection:conditional}

Researchers aim for ``plug-and-play'' (PnP) frameworks \citep{dathathri_plug_2020} which can be used along an existing generative LM (referred to as the ``base LM'') with minimum or no interference between the PnP components and the base LM.

Comparing to non-plug-and-play methods ("white-box" approaches),
these frameworks can be roughly classified into three categories. 
{\em Gray-box} approaches access and modify some non-input-output layer computations, usually the hidden representation, hence ``plugging'' an additional model in the middle of the base LM~\citep{dathathri_plug_2020,madotto_plug-and-play_2020,duan_pre-train_2020,mai_plug_2020}.
{\em Black-box} approaches including ``Prompt Engineering'' that aim to change the prompts fed into the base LM at inference time~\citep{wallace_universal_2019,li_prefix-tuning_2021}. 
{\em Guided generation} targets at building a controllable ``guiding'' model that shifts the output from base LM at inference time~\citep{krause_gedi_2020,pascual_directed_2020,liu_--fly_2021}.

The generation model we propose is an extension of GeDi~\citep{krause_gedi_2020}. 
We enhanced it with additional capabilities to support multi-topic generation with continuous weighting, supporting the downstreaming applications while keeping its capability to transfer to different base LMs.
\subsubsection{Controllable Story Generation}
Procedural generation of stories has been studied for more than 40 years, and researcher has attempted to generate stories with constrained plots and entities that plot acts upon \cite{kybartas_survey_2017}.
We contextualize our work in the field of \textit{neural} controllable story generation, where new stories are generated from imitating existing stories.
Neural story generation systems do this by training or fine-tuning a language model on story data.
Sampling from a language model trained on story data tends to result in text output that looks like stories as well.
However, naively sampling from 
$P_{\theta}(x_t | x_{<t})$ (See Section \ref{section_ppglm}) or from any other neural systems
is uncontrolled in the sense that one does not have any influence over the output after the initial context input.



A number of neural story generation systems have attempted to condition the generation with some form of high-level plan.
Storytelling systems such as \citep{akoury-etal-2020-storium, yao_plan-and-write_2019} embeds topic constraints directly into the model.
These systems extract a set of topics from a dataset that must be incorporated into the story.
\textit{PlotMachines} \citep{rashkin-etal-2020-plotmachines} 
allows a human user to specify topics that can be incorporated into a story in any order.
\citet{wang_narrative_2020} generate a story by interpolating between a start event and an end event in a slot filling fashion, targeted the same goal.
Our work differs in two ways.
First, we allow blending of topics such that a single line in a story can meet more than one topic provided by a human user.
Second, we have developed a black-box plug-and-play system that works with different LMs.





\section{Preliminaries}
\label{section_ppglm}
Generative Language Models (LMs), specifically continuation models, take a context (``prompt'') and generate a continuation by predicting the next tokens.
This is achieved by optimizing the model parameters $\theta$ that best estimates the probability density of a sequence of word tokens 
$x_{1: T}=\left\{x_{1}, \ldots, x_{T}\right\}$
represented as an auto-regressive factorization:
\begin{equation}
P_{\theta}\left(x_{1: T}\right)=\prod_{t=1}^{T} P_{\theta}\left(x_{t} \mid x_{<t}\right).
\label{eq_bg_1}
\end{equation}
By iteratively predicting a distribution on the next token given the previous tokens, a continuation can be generated by repeatedly sampling $P_{\theta}\left(x_{t} \mid x_{<t}\right)$ and attach the selected token back to the ``previous'' tokens for the next step.

Sequences generated this way are uncontrolled;
To control the generated sequence, an \textbf{attribute} represented as a class variable \cite{keskar_ctrl_2019} that could describe sentiment, topics or anything that fit into a class, can be introduced to equation \eqref{eq_bg_1} to form a Class-Conditional Language Model (CC-LM):
\begin{equation}
P_{\theta}\left(x_{1: T} \mid c\right)=\prod_{t=1}^{T} P_{\theta}\left(x_{t} \mid x_{<t}, c\right)
\label{eq_bg_2}
\end{equation}
where $c$ represents the class variable, or ``control code'', that describes an \textbf{attribute} of the sequence $x_{1: T}$.
However, since $c$ and $x_{1: T}$ are entangled in equation \eqref{eq_bg_2}, naively optimizing $P_{\theta}$ requires a new CC-LM to be trained.

To decouple the conditional generation component, $c$, from the unconditional part, $P_{L M}\left(x_{1: T}\right)$, \cite{krause_gedi_2020} proposed the GeDi framework and an algorithm to enable a separate controlling model to guide the generation process of a base language model.
Instead of tackling $P_{\theta}\left(x_{1: T} \mid c\right)$ directly, they train a contrastive discriminator model on the side to estimate
\begin{equation}
\begin{aligned}
   P_{\theta}\left(c \mid x_{1: t}\right) = 
   \alpha P(c) \prod_{j=1}^{t} P_{\theta}\left(x_{j} \mid x_{<j}, c\right)
\end{aligned}
\end{equation}
where 
$\alpha$ is the normalization constant $\alpha = 1/ (\sum_{c^{\prime} \in\{c, \bar{c}\}} \prod_{j=1}^{t} P\left(c^{\prime}\right) P_{\theta}\left(x_{j} \mid x_{<j}, c^{\prime}\right))$, and
$c$ and $c^{\prime}$ are contrastive control codes ($c$ and not-$c$).
At the decoding stage of the generation process, one can guide the generation by using $P_{\theta}\left(c \mid x_{1: t}\right)$ as a posterior to the output probability distribution of the base LM:
\begin{equation}
\begin{aligned}
   P&\left(x_{t} \mid x_{<t}, c\right) \propto \\ 
   & P_{L M}\left(x_{t} \mid x_{<t}\right) P_{\theta}\left(c \mid x_{t}, x_{<t}\right)^{\omega}
\end{aligned}
\label{eq_bg_4}
\end{equation}
where $\omega$ is a parameter for control strength, with larger values biasing generation more strongly towards $c$.
CC-LMs trained this way do not require access to any internal data of the base LM, and works independently of it.


\section{The Plug-and-Blend Framework}

Our {\em Plug-and-Blend} framework consists of two components (figure \ref{fig:overall}):
(1)~a {\em blending generative Model} (BGM) that is responsible for plug-and-play controlled continuations using the control specifications; and 
(2)~a {\em planner} that plans and assigns control specifications based on Control Sketches.

A {\em Control Sketch} is a high-level specification of what topics should be present in the story and what portions of the story each topic should approximately appear in. 
This provides a human co-creator the ability to guide the generator loosely, with a broad range per topic, or tightly, with a narrow range per topic.
We envision a co-creative loop wherein the human user provides a control sketch and iteratively updates the control sketch based on generation results, refining the topics and refining the ranges for the topics.
The user interface for eliciting control sketches from a human is outside the scope of this paper and experiments about the co-creative loop are left for future work. 
The next sections provide the algorithmic support for control sketches.


\subsection{Blending Generative Model}
\label{section:generation_model}

The BGM generates the sentence continuation. It consists of two parts, a (1)~plug-and-play language model and (2)~a control model. 
Given a prompt $x_{<t}$, the plug-and-play language model produces a vector of logits $P_{L M}\left(x_{t} \mid x_{<t}\right)$. 
The control model biases the output of the language model toward particular tokens associated with the topics of the control codes ${c \in C}$ based on the desired strengths of each topic $
\omega_{{c \in C}}^{*} \in \Omega$.
Together the two models iteratively find the best token $x_{t}$ that reflects both natural language composition
and control bias presented by $c$ and $\omega$.
A larger $\omega_{{c}}^{*}$ means more steering towards the topic represented by control code $c$.

Inspired by the application of generative adversarial networks to latent space walking, we treat $P_{\theta}\left(c \mid x_{t}, x_{<t}\right)$ (described in section \ref{section_ppglm}) as a heuristic of \textbf{direction} that increases $P\left(x_{t} \mid x_{<t}, c\right)$ in a $|V|$-dimensional latent space, where $V$ is the language model's vocabulary.
For example, consider two different control codes $c_1$ and $c_2$ instantiating equation \eqref{eq_bg_4}.
To apply both control codes in the generation process, we use the heuristic 
\begin{equation}
\begin{aligned}
    P&\left(x_{t} \mid x_{<t}, c_1, c_2\right) \propto 
   P_{L M}\left(x_{t} \mid x_{<t}\right) \times \\
   & P_{\theta}\left(c_1 \mid x_{t}, x_{<t}\right)^{\omega_1}
   P_{\theta}\left(c_2 \mid x_{t}, x_{<t}\right)^{\omega_2}
\end{aligned}
\end{equation}
to combine the effect of both posterior distributions into one universal posterior.
$\omega_1$ and $\omega_2$ in this case represents control strength for each control code, $c_1$ and $c_2$ respectively, and can be different, enabling continuous blending between topics. 
This process can be repeated with a set of control codes $C = \left\{c_{1}, \ldots, c_{n}\right\}$ with weights $\Omega = \left\{\omega_{1}, \ldots, \omega_{n}\right\}$.

Formally, at the decoding stage of the generation process, a control model compute controlled probability using the following equation:
\begin{equation}
\begin{aligned}
   P&\left(x_{t} \mid x_{<t}, C\right) = \\
   & P_{L M}\left(x_{t} \mid x_{<t}\right) \prod_{c^{*} \in C}  P_{\theta}\left(c^{*} \mid x_{t}, x_{<t}\right)^{\omega_c^{*}}
\end{aligned}
\label{eq_bg_6}
\end{equation}
where the control strengths of individual control codes are normalized with
$\sum_{c} \omega_c^{*}=\omega$, where $\omega$ is total control strength.\footnote{This is not the only way to formalize this heuristic; We found this to be effective and efficient.}
This can be efficiently computed by batching input sequences appended by different control codes, with little overhead compared to the original GeDi~\citep{krause_gedi_2020} framework. 


\subsection{Planner}
\label{section:planner}

The human user provides a high-level control sketch of the story, consisting of the number of sentences, $N$, a set of topics, $C$, and a range of lines to which to apply the topic, $r:=(s,e)$ where $s \leq e$.

Formally, we denote planner setup as $(N,SK)$ where $SK:={sk_1,\ldots,sk_m}$, $sk:=(c,r)$ and $c \in C$.
Sketches can have their range $r$ overlap such that multiple topics can be applied to the same lines of the story.

Given the control sketch, the planner produces a control configuration $C_n, \Omega_n$ for each sentence position $n = \{0, \ldots, N-1\}$.
The control configuration for each sentence is passed to BGM,
along with previous generated sentences as prompt.

We interpret a control sketch as a story arc on a specific topic, which typically contains a transition, an engagement, and a phase-out,
the planner should give the highest control strength to the midpoint of the area, $m := (s+e)/2$, and lower strength towards the start and end of the span of the area;
We capture this as a Gaussian distribution.

Formally, the following equation translates the sketch into a control configuration for each position $n \in N$:
\begin{equation}
\label{eqn:pdf}
    \omega^{+}_{c,n} = f(\mathcal{N}(m,(\sigma / (e-s+\epsilon)^{2}))(n-m)
\end{equation}
where $f(\cdot)$ indicates probability density function, $\epsilon$ is an infinitesimal,
and $\sigma$ is a tunable parameter representing overall transition smoothness, where higher $\sigma$ grants smoother transitions in the cost of reduced topic engagement for midpoint. 
Since there can be multiple control sketches and they can be of the same control code, we apply each individual sketch in the order they are presented and normalize after each application so that $\Sigma_{n} \omega_{c,n} = 1$.
Table \ref{table_planner_example} shows 2 example sketches and their corresponding generated stories by this planner.




\section{Blending Generation Experiments}

For evaluating BGM, we use the GPT2-large model fine-tuned on ROCStories~\cite{mostafazadeh_corpus_2016} as our base language model.
Fine-tuning GPT2 on ROCStories results in a model that generates short stories about common everyday situations.
We pair the language model with a pre-trained GeDi (which in turn is based on GPT2-medium) trained on AG-news\footnote{\url{http://groups.di.unipi.it/~gulli/AG_corpus_of_news_articles.html}} as the guiding model.
Across all setups, at generation time, we use greedy decoding with repetition penalty described in \citet{keskar_ctrl_2019}, and only use the first sentence generated as the output, discarding any token after it.


Since there is no ground truth for any generated sequence, metrics such as BLEU and other n-gram-based metrics are not applicable.
This poses a unique challenge in evaluating our system, limiting us to unsupervised metrics.
In this section, we report evaluation of our blending generative model from two aspects: 
\begin{itemize}
    \item Fluency: measuring how our generated sequence forms natural language; and
    \item Control fidelity: measuring how our generated sequence respects the requested control codes and strength.
\end{itemize}


\subsection{Blending Fluency}


To evaluate fluency of sequences generated by our BGM, we use perplexity of {\em base} language model;
perplexity is highly correlated with fluent generation in LMs.
The intuition is that if generated sentences have low average perplexity when evaluated by the base LM then they are consistent with sentences we would find in the base LM's training data.

To generate sequences from our model, we used 100 sentences from a 10\% held-out evaluation set of ROCStories not seen at fine-tuning time. ROCStories contains five-sentence stories; we always pick the first sentence. 
That sentence becomes our prompt and is paired with all possible combinations of two topic choices chosen from ``Business'', ``Science'', ``Sports'', or ``World''.
These are the topics that the GeDi model is optimized for.
Our control sketch gives equal blending weighting for all topics.
We vary the control strength using the following increments: $[0, 0.5, 1, 1.5, 2, 3, 4]x$, where 0 represents an uncontrolled base LM and $4x$ represents 400\% of the control strength hyperparameter used by \citet{krause_gedi_2020}.

\begin{figure}
    \centering

    \includegraphics[width=1\columnwidth]{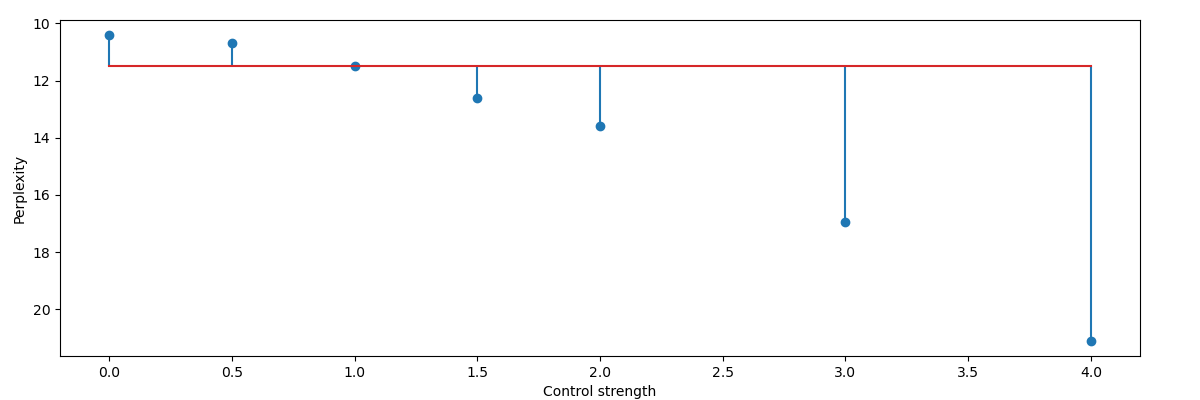}

    \caption{Perplexity (lower is better) of generated sequences with 2 topics. Baseline performance (red line) set at $1x$.
    }
    \label{fig:ppl_roc}
\end{figure}

Figure \ref{fig:ppl_roc} shows the average perplexity of generated sequences, measured by the Base LM.
We observe that average perplexity increases with stronger control, signaling a departure of generated sequences from what the base LM would generate, and a potential decrease in fluency.
This is to be expected as the control is biasing the generated text more and more toward the use of words that are consistent with a particular topic and away from general word frequency.
In the region of 0 to $2x$ control strength, we see minor and more linear increase of perplexity compared to $>2x$ region.

\begin{table*}[t]
\footnotesize
\begin{tabular}{l} 
{\bf Prompt:} The people gathered to protest the court's ruling last week.
\end{tabular}

\begin{tabular}{|c|c|m{0.5\textwidth}||c|c|}
\hline
\bm{$c_1=$} {\bf Sports} & \bm{$c_2=$} {\bf Business} & \multirow{2}{*}{\bf BGM-Generated Sentence (Overall Strength 2x)} & \multicolumn{2}{c|}{\bf Classifier score}\\
\bm{$\omega_{c_1}$} & \bm{$\omega_{c_2}$} &  & \bm{$c_1$} & \bm{$c_2$} \\
\hline
100\% & 0\% & Coach Leeman was in a wheelchair and had been taken to hospital for treatment. & 86\%& 14\% \\
\hline
75\% & 25\% & Coach Reebok was one of them. & 65\% & 35\%  \\
\hline
50\%  & 50\% & The players were joined by a few of them. & 84\% & 16\% \\
\hline
25\%  & 75\% & The company that owns the team was fined \$1,000 for violating a rule prohibiting employees from using their own equipment. & 37\% &63\% \\
\hline
0\% & 100\% & Bankruptcy Judge William H. said that the bank had failed to pay its creditors and was in default on \$1 billion of loans it owed them. & 24\% & 76\% \\
\hline
\end{tabular}

\begin{tabular}{c} 
Comparing column 1 with column 4, Kendall's $\tau$-a $= 0.8$ for this generated sequence.
\end{tabular}

\caption{Example evaluation of control fidelity. The first two columns indicate the requested control strengths. The last two columns indicate the probability that each line is either Sports or Business based on a BART-based topic classifier. We expect to see the classifier score for $c_1$ decrease as the classifier score for $c_2$ increases.}
\label{table_fidelity_example}
\end{table*}

\subsection{Control Fidelity}
\label{exp_2}

Control fidelity is how well the generator responds to multiple control codes applied at once (see \citet{krause_gedi_2020} for experiments applying one control code at a time; we do not replicate them in this paper).
For story generation, multiple control codes can be applied to the same sentence in a story at different weights.
We perform experiments in a latent space walking setting, measuring content changes of generated sentences under the same prompt and control codes but different relative control strength.

Given a particular prompt line in a story and two control topics $c_1$ and $c_2$, we re-generate the same line multiple times under different control strengths for each topic. 
Specifically we set $\omega_{c_1}$ to 0\%, 25\%, 50\%, 75\% or 100\% and $\omega_{c_2}=1-\omega_{c_1}$ to represent a range of different possible blends of topics in the same line.
See Table \ref{table_fidelity_example} for an example.
Since we know the control parameters used to generate these sentences, in which $c_1$ receives more and more control strength relative to $c_2$, we expect to see sentences that are increasingly about topic $c_1$ and decreasingly about topic $c_2$. 
These sentences do not comprise a story sequence, but are different alternative sentences for the same line in a story under different topic control specifications.



To determine whether a given generated sentence was representative of a topic, we score each generated sentence with an off-the-shelf BART-based zero-shot classifier\footnote{huggingface pipeline("zero-shot-classifier")}
with $c_1$ and $c_2$, in raw text form, as possible classes.
We then compare the order of the sentences as determined by the classifier to the ground-truth order of increasing control strength of $c_1$.
We report the correlation of order between these two sequences using Kendall's $\tau$-a metric.
A perfectly strictly increasing classifier score will grant a $\tau$-a score of $1$ for a sequence.
If the sentences have some reordering based on classification score, $\tau$-a is reduced.
A score of 0 indicates a random ordering and and a score of $-1$ indicates a sequence that is exactly in opposite order.
Table~\ref{table_fidelity_example} shows the classifier scores for the possible next sentences under different control strengths; the classifier scores are not monotonically decreasing, resulting in a $\tau$-a score of 0.8.

\begin{figure*}[h!]
    \centering
    \begin{subfigure}[h]{0.5\columnwidth}
        \includegraphics[width=1\columnwidth]{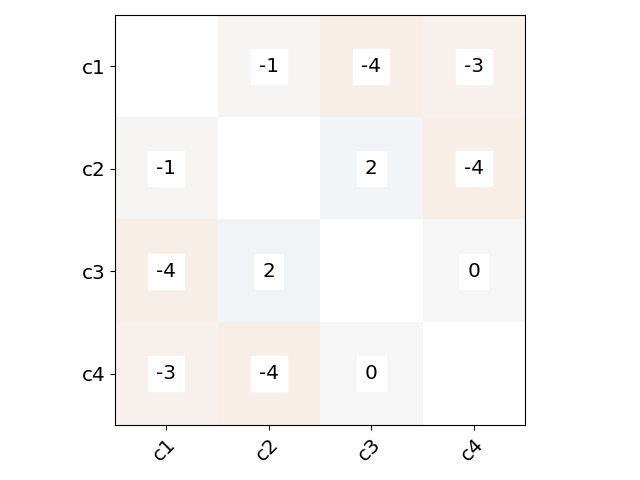}
        \caption{Baseline.}
        \label{subfig:baseline}
    \end{subfigure}%
    \begin{subfigure}[h]{0.5\columnwidth}
        \includegraphics[width=1\columnwidth]{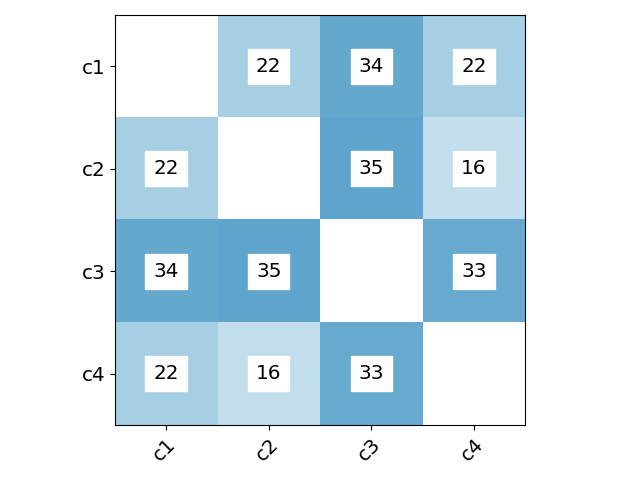}
        \caption{Total control strength $1x$.}
    \end{subfigure}
        \begin{subfigure}[h]{0.5\columnwidth}
        \includegraphics[width=1\columnwidth]{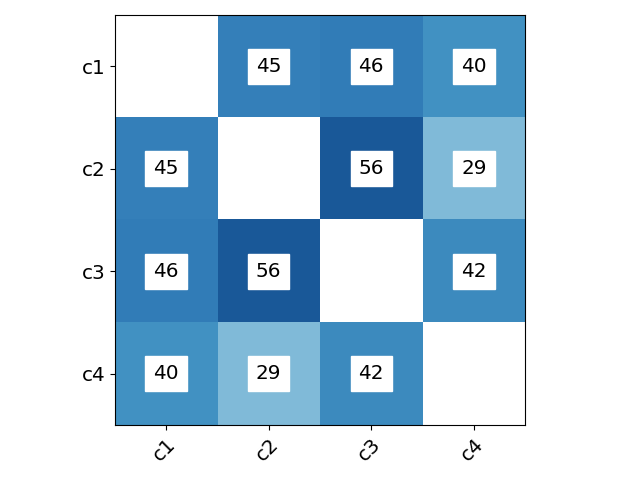}
        \caption{Total control strength $2x$.}
    \end{subfigure}%
        \begin{subfigure}[h]{0.5\columnwidth}
        \includegraphics[width=1\columnwidth]{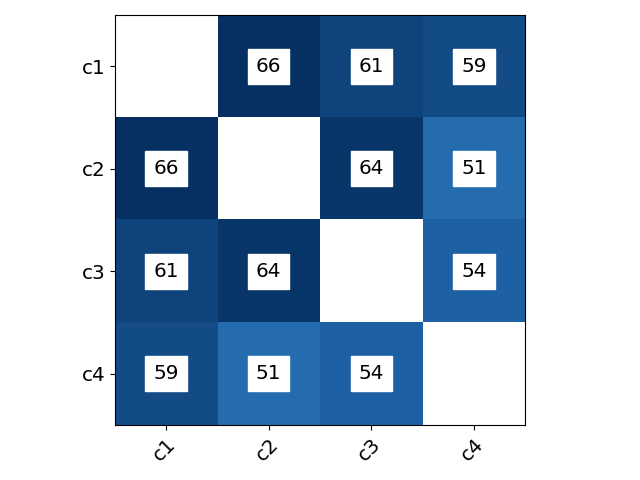}
        \caption{Total control strength $4x$.}
    \end{subfigure}
    \caption{average $\tau$-a (higher meaning better control fidelity) under various Total control strength for the tuned model with topics: (c1) Business, (c2) Science, (c3) Sports, (c4) World, comparing to baseline, in percentages ($-100\%\dots 100\%$). }
    \label{fig:tau_tuned}
\end{figure*}

Figure \ref{fig:tau_tuned} shows a heat-map of the average $\tau$-a score of sequences of sentences generated with different control code pairs and different total control strength (percentages).
For each combination of parameters, 100 sequences of 5 sentences are generated and evaluated.
We focus on comparisons instead of absolute values of the $\tau$-a metrics, as the classifier we used introduce noise to the metrics. 
Comparing to the baseline, which is the evaluation metric applied to order-randomized stories in ROCStories dataset, we observe universal statistical significance ($p < .01$) in improvement in $\tau$-a metric. 
That is, without a control bias, rank ordering is random.
As we increase the total control strength, the rank order of generated sentences more closely matches the ground truth order.

Some topic combinations (e.g., Science and Sports) work better than others (e.g., Science and World); the ``World'' category appears to include a lot of overlapping vocabulary usage with the other categories.
From these results, we show the plug-and-blend technique 
(a)~significantly increases the likelihood that topics will be incorporated into sentences, and
(b)~is sensitive to blended topics.



\begin{figure*}[h!]
    \centering
    \begin{subfigure}[h]{0.5\columnwidth}
        \includegraphics[width=1\columnwidth]{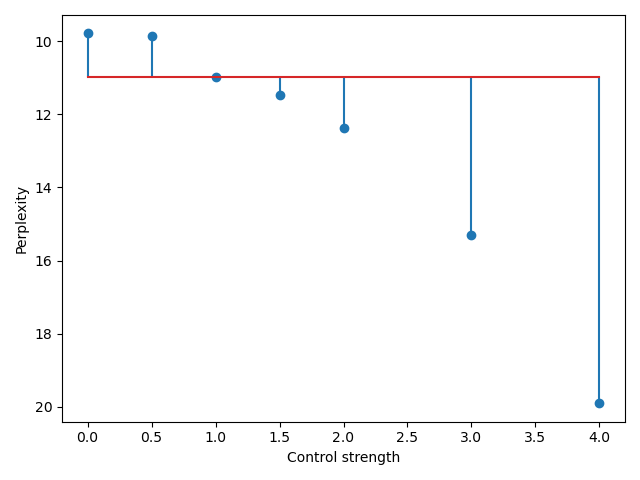}
        \caption{Perplexity of generated sequences.}
    \end{subfigure}%
    \begin{subfigure}[h]{0.5\columnwidth}
        \includegraphics[width=1\columnwidth]{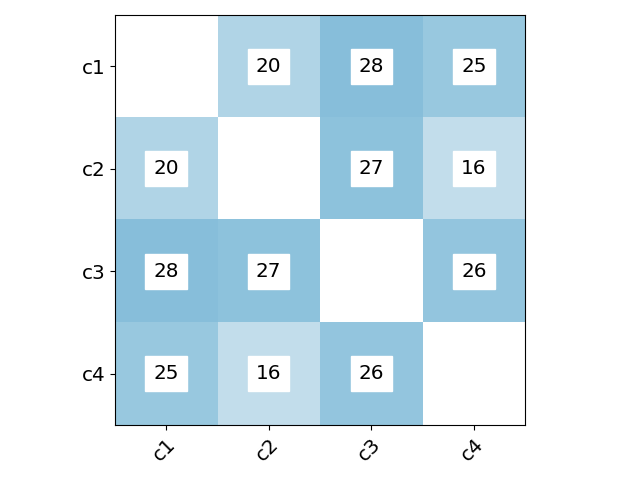}
        \caption{Total control strength 1x.}
    \end{subfigure}
        \begin{subfigure}[h]{0.5\columnwidth}
        \includegraphics[width=1\columnwidth]{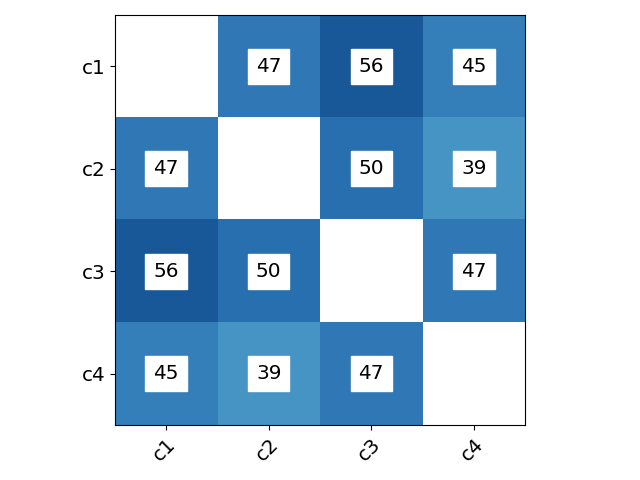}
        \caption{Total control strength 2x.}
    \end{subfigure}%
        \begin{subfigure}[h]{0.5\columnwidth}
        \includegraphics[width=1\columnwidth]{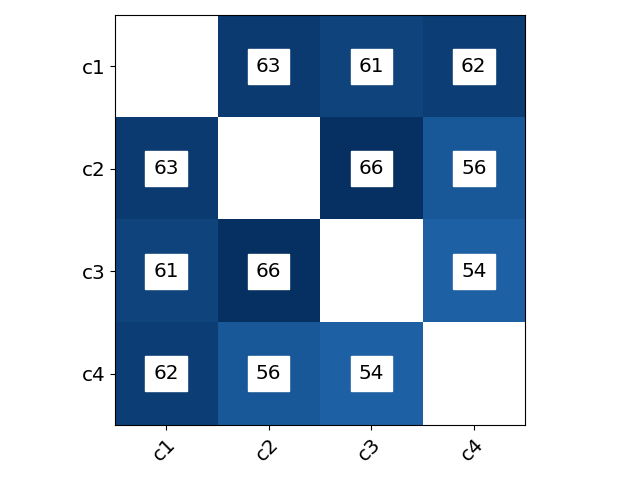}
        \caption{Total control strength 4x.}
    \end{subfigure}
    \caption{Experiment results for the untuned model. Refer to Figure \ref{subfig:baseline} for baseline comparison.}
    \label{fig:tau_untuned}
\end{figure*}

Figure~\ref{fig:tau_untuned} shows the same experiment as above, but with a non-fine-tuned version of GPT2-large.
This shows that the plug-and-blend technique works on language models that haven't been fine-tuned on ROCStories.
The prompts are still from ROCStories, but are not as representative of the untuned model. 
In this condition, the text generated will not read as sentences in stories.
We observe similar improvements over the baseline, demonstrating the ability of our method in keeping the strong adaptation capability.



\section{Planner Evaluation}

\begin{small}
\begin{table*}[h!]
\centering
\scriptsize
\begin{tabular}{|c|p{0.75\textwidth}|}
\hline
{\bf Control Sketches} & {\bf Generated Paragraph} \\ \hline
\begin{tabular}[c]{@{}c@{}}{\bf Sketch 1}\\ $c_1={\rm Sports}$, \{$0\rightarrow 5$\}\\ $c_2={\rm Science}$, \{\textbf{4} $\rightarrow 10$\}\end{tabular} & \begin{tabular}[c]{@{}p{0.75\textwidth}@{}}Jackie Robinson was playing in the NBA.\\ He had just been drafted by the Cleveland Cavaliers.\\ He decided to play for a team that was trying to win a championship.\\ "I'm gonna go out there and beat them," he said "They're going to be great."\\ "Yeah, they're good," he said "they'll be great."\\ He added that his wife had been working on the project for about a year.\\ He was happy to see her progress.\\ She was able to get some work done and she started getting more comfortable with it.\\ She had a lot of fun doing it!\\ She also got her own computer, which is now in the living room.\end{tabular} \\ \hline
\begin{tabular}[c]{@{}c@{}}{\bf Sketch 2}\\ $c_1={\rm Sports}$, $0\rightarrow 5$\\ $c_2={\rm Science}$, \textbf{5} $\rightarrow 10$\end{tabular} & \begin{tabular}[c]{@{}p{0.75\textwidth}@{}}Jackie Robinson was playing in the NBA.\\ He had just played basketball for a few years.\\ He was looking forward to his first game of the season.\\ He decided to play with his friends and play against them in the court.\\ He got a lot of feedback from everyone who played against him, including some that were very excited about it!\\ I was really happy when I saw how he played.\\ I also had to admit that my favorite player was the guy who beat me in the finals.\\ The computer game Super Mario 64 is a great game, but it's not perfect.\\ I played it on my laptop and found that I couldn't play it properly because of some bugs.\\ The problem was that the graphics were bad, so I had to use an emulator instead of playing the game.\end{tabular} \\ \hline

\end{tabular}
\caption{Generated Examples with different Control-Sketches.}
\label{table_planner_example}
\end{table*}
\end{small}

We now investigate the planner on how well it utilizes BGM in generating stories respecting human-provided sketches, by
conducting human evaluations on stories generated.



\subsubsection{Experiment setup}
We selected the first 5 first-sentence-of-the-story from the held-out evaluation set of ROCStories, and request our planner, working with the tuned BGM to generate stories with two sketch setup:
\textbf{(1) Story with Transitions ("Ours")} use a sketch requiring 10 lines and two categories, $c_1$ and $c_2$ taken from the four topics used in the previous section. 
$c_1$ applied to the first 5 lines and $c_2$ the last 5 lines. 
This setup demonstrates an intent to generate a story with transition between two topics;
\textbf{(2) No Blending Baseline} uses a sketch requiring 10 lines and a single category $c$ applied to every line. 
This setup effectively disables both BGM and the planner and serves as a baseline intent of a story concentrating on a single topic.

We recruited 49 
participants on Prolific \footnote{www.prolific.co}.
Each participant  
read 7 random pairs of stories from both setups that were generated from the same prompt and same initial topic, and additional training and attention check pairs.
We ask participants to choose which story contains a transition between two topics supplied (or the one containing a stronger transition if the participant believes both have a transition) and provide their reasoning.\footnote{The reasoning text entry is strictly to provide evidence of attention to the task and we do not analyze this data further.}
We accepted 45 submissions which passed our attention check.
We additionally provide a ``Neither'' option if participants do not sense a transition in either of the two stories presented.
However, in a control experiment where the same participants read two stories both generated by the baseline, 51\% ($n=45$) of participants find a transition, 
prompting us to not use
this feature for metrics on detecting a transition for evaluating our approach.
We {\em do not} ask participants to choose which story they like better because the language model is the same in both conditions and the objective is to determine if participants can detect the topic transitions, indicating that users will be able to affect control.

For the main experiment, we collected ($n=201$) relevant answers from the 45 submissions for evaluation.

\subsubsection{Results}

Participants choose the stories generated by ``Ours'' 75.1\% ($n=201$) of the time, which is significantly more likely ($p < 0.01$, post-hoc power $>99\%$) than the baseline.
This preference of finding a topic transition strongly support the effectiveness of our planner on injecting transitions into generating stories when requested.
We conclude that our planner respects sketch input and translates transition requests from the sketches to transition signals in the stories generated, and that these transitions are perceptible by human readers. Table~\ref{table_planner_example} shows two example output stories and their sketches.

\section{Discussion}


For BGM, there is a trade-off between control fidelity and fluency.
%
In practice, an iterative deepening algorithm can be used where
increasing Control strength modifiers can be tried and the best generated sentence, as measured by task-specific metric such as \cite{guan-huang-2020-union}), is selected.
This can, just like how multiple control codes are handled, be implemented very efficiently.

The current planner is heuristic. 
We envision a planner that can be parameterized and learn from demonstrations.
Reinforcement learning, in which the context and control sketches work as world states, can choose control configurations as actions.
Feedback (reward) from the creator would be necessary.
This would incorporate the plug-and-blend technique into a co-creative process wherein the generator learns to blend preferences from the human creator ~\citep{guzdial_co-creative_2018}.

In this paper, we have made no assumptions about how the planner acquire control sketches from the human content creator.
We envision a process whereby the human co-creator has an idea of what the final product---in this case a story---should be and can express that vision at the level of how topics manifest through the story.
The control model will need to support different topics at different levels of granularity; currently the control model only supports four topics, which is sufficient for conducting experiments to characterize the plug-and-blend technique but not for full co-creativity.


\section{Conclusions}
In this paper, we present Plug-and-Blend, a plug-and-play framework that enhances a base LM, enables controllable generation with continuous-weighted control codes, along with capability of generating stories based on human intentions, all without access to internal knowledge of this base model.
These capabilities will fuel a new generation of PCG applications with the key assets of high-level controls available to content creators, decoupling between the controllable component and the generative component, and easiness of adapting to new advancements in the field of generative models.


\bibliography{anthology,references}

\begin{thebibliography}{21}
\providecommand{\natexlab}[1]{#1}
\providecommand{\url}[1]{\texttt{#1}}
\providecommand{\urlprefix}{URL }
\expandafter\ifx\csname urlstyle\endcsname\relax
  \providecommand{\doi}[1]{doi:\discretionary{}{}{}#1}\else
  \providecommand{\doi}{doi:\discretionary{}{}{}\begingroup
  \urlstyle{rm}\Url}\fi

\bibitem[{Akoury et~al.(2020)Akoury, Wang, Whiting, Hood, Peng, and
  Iyyer}]{akoury-etal-2020-storium}
Akoury, N.; Wang, S.; Whiting, J.; Hood, S.; Peng, N.; and Iyyer, M. 2020.
\newblock {STORIUM}: {A} {D}ataset and {E}valuation {P}latform for
  {M}achine-in-the-{L}oop {S}tory {G}eneration.
\newblock In \emph{Proceedings of the 2020 Conference on Empirical Methods in
  Natural Language Processing (EMNLP)}, 6470--6484. Online: Association for
  Computational Linguistics.
\newblock \doi{10.18653/v1/2020.emnlp-main.525}.
\newblock \urlprefix\url{https://www.aclweb.org/anthology/2020.emnlp-main.525}.

\bibitem[{Brown et~al.(2020)Brown, Mann, Ryder, Subbiah, Kaplan, Dhariwal,
  Neelakantan, Shyam, Sastry, Askell, Agarwal, Herbert-Voss, Krueger, Henighan,
  Child, Ramesh, Ziegler, Wu, Winter, Hesse, Chen, Sigler, Litwin, Gray, Chess,
  Clark, Berner, McCandlish, Radford, Sutskever, and
  Amodei}]{brown_language_2020}
Brown, T.~B.; Mann, B.; Ryder, N.; Subbiah, M.; Kaplan, J.; Dhariwal, P.;
  Neelakantan, A.; Shyam, P.; Sastry, G.; Askell, A.; Agarwal, S.;
  Herbert-Voss, A.; Krueger, G.; Henighan, T.; Child, R.; Ramesh, A.; Ziegler,
  D.~M.; Wu, J.; Winter, C.; Hesse, C.; Chen, M.; Sigler, E.; Litwin, M.; Gray,
  S.; Chess, B.; Clark, J.; Berner, C.; McCandlish, S.; Radford, A.; Sutskever,
  I.; and Amodei, D. 2020.
\newblock Language {Models} are {Few}-{Shot} {Learners}
  \urlprefix\url{https://arxiv.org/abs/2005.14165v4}.

\bibitem[{Dathathri et~al.(2020)Dathathri, Madotto, Lan, Hung, Frank, Molino,
  Yosinski, and Liu}]{dathathri_plug_2020}
Dathathri, S.; Madotto, A.; Lan, J.; Hung, J.; Frank, E.; Molino, P.; Yosinski,
  J.; and Liu, R. 2020.
\newblock Plug and {Play} {Language} {Models}: {A} {Simple} {Approach} to
  {Controlled} {Text} {Generation}.
\newblock \emph{International Conference on Learning Representations} (2020).
\newblock \urlprefix\url{http://arxiv.org/abs/1912.02164}.
\newblock ArXiv: 1912.02164.

\bibitem[{Duan et~al.(2020)Duan, Xu, Pei, Han, and Li}]{duan_pre-train_2020}
Duan, Y.; Xu, C.; Pei, J.; Han, J.; and Li, C. 2020.
\newblock Pre-train and {Plug}-in: {Flexible} {Conditional} {Text} {Generation}
  with {Variational} {Auto}-{Encoders}.
\newblock \emph{Proceedings of the 58th Annual Meeting of the Association for
  Computational Linguistics} (2020): 253--262.
\newblock \urlprefix\url{http://arxiv.org/abs/1911.03882}.
\newblock ArXiv: 1911.03882.

\bibitem[{Fang et~al.(2021)Fang, Zeng, Liu, Bo, Dong, and
  Chen}]{fang_transformer-based_2021}
Fang, L.; Zeng, T.; Liu, C.; Bo, L.; Dong, W.; and Chen, C. 2021.
\newblock Transformer-based {Conditional} {Variational} {Autoencoder} for
  {Controllable} {Story} {Generation}.
\newblock \emph{arXiv:2101.00828 [cs]}
  \urlprefix\url{http://arxiv.org/abs/2101.00828}.
\newblock ArXiv: 2101.00828.

\bibitem[{Guan and Huang(2020)}]{guan-huang-2020-union}
Guan, J.; and Huang, M. 2020.
\newblock {UNION}: {A}n {U}nreferenced {M}etric for {E}valuating {O}pen-ended
  {S}tory {G}eneration.
\newblock In \emph{Proceedings of the 2020 Conference on Empirical Methods in
  Natural Language Processing (EMNLP)}, 9157--9166. Online: Association for
  Computational Linguistics.
\newblock \doi{10.18653/v1/2020.emnlp-main.736}.
\newblock \urlprefix\url{https://www.aclweb.org/anthology/2020.emnlp-main.736}.

\bibitem[{Guzdial, Liao, and Riedl(2018)}]{guzdial_co-creative_2018}
Guzdial, M.; Liao, N.; and Riedl, M. 2018.
\newblock Co-{Creative} {Level} {Design} via {Machine} {Learning}.
\newblock \emph{Fifth Experimental AI in Games Workshop}
  \urlprefix\url{http://arxiv.org/abs/1809.09420}.
\newblock ArXiv: 1809.09420.

\bibitem[{Hu et~al.(2017)Hu, Yang, Liang, Salakhutdinov, and
  Xing}]{hu_toward_2017}
Hu, Z.; Yang, Z.; Liang, X.; Salakhutdinov, R.; and Xing, E.~P. 2017.
\newblock Toward {Controlled} {Generation} of {Text}.
\newblock In Precup, D.; and Teh, Y.~W., eds., \emph{Proceedings of the 34th
  {International} {Conference} on {Machine} {Learning}}, volume~70 of
  \emph{Proceedings of {Machine} {Learning} {Research}}, 1587--1596.
  International Convention Centre, Sydney, Australia: PMLR.
\newblock \urlprefix\url{http://proceedings.mlr.press/v70/hu17e.html}.

\bibitem[{Keskar et~al.(2019)Keskar, McCann, Varshney, Xiong, and
  Socher}]{keskar_ctrl_2019}
Keskar, N.~S.; McCann, B.; Varshney, L.~R.; Xiong, C.; and Socher, R. 2019.
\newblock {CTRL}: {A} {Conditional} {Transformer} {Language} {Model} for
  {Controllable} {Generation}.
\newblock \emph{arXiv:1909.05858 [cs]}
  \urlprefix\url{http://arxiv.org/abs/1909.05858}.
\newblock ArXiv: 1909.05858.

\bibitem[{Krause et~al.(2020)Krause, Gotmare, McCann, Keskar, Joty, Socher, and
  Rajani}]{krause_gedi_2020}
Krause, B.; Gotmare, A.~D.; McCann, B.; Keskar, N.~S.; Joty, S.; Socher, R.;
  and Rajani, N.~F. 2020.
\newblock {GeDi}: {Generative} {Discriminator} {Guided} {Sequence}
  {Generation}.
\newblock \emph{arXiv:2009.06367 [cs]}
  \urlprefix\url{http://arxiv.org/abs/2009.06367}.
\newblock ArXiv: 2009.06367.

\bibitem[{Kybartas and Bidarra(2017)}]{kybartas_survey_2017}
Kybartas, B.; and Bidarra, R. 2017.
\newblock A {Survey} on {Story} {Generation} {Techniques} for {Authoring}
  {Computational} {Narratives}.
\newblock \emph{IEEE Transactions on Computational Intelligence and AI in
  Games} 9(3): 239--253.
\newblock ISSN 1943-0698.
\newblock \doi{10.1109/TCIAIG.2016.2546063}.
\newblock Conference Name: IEEE Transactions on Computational Intelligence and
  AI in Games.

\bibitem[{Li and Liang(2021)}]{li_prefix-tuning_2021}
Li, X.~L.; and Liang, P. 2021.
\newblock Prefix-{Tuning}: {Optimizing} {Continuous} {Prompts} for
  {Generation}.
\newblock \emph{arXiv:2101.00190 [cs]}
  \urlprefix\url{http://arxiv.org/abs/2101.00190}.
\newblock ArXiv: 2101.00190.

\bibitem[{Liu et~al.(2021)Liu, Sap, Lu, Swayamdipta, Bhagavatula, Smith, and
  Choi}]{liu_--fly_2021}
Liu, A.; Sap, M.; Lu, X.; Swayamdipta, S.; Bhagavatula, C.; Smith, N.~A.; and
  Choi, Y. 2021.
\newblock On-the-{Fly} {Controlled} {Text} {Generation} with {Experts} and
  {Anti}-{Experts}.
\newblock \emph{arXiv:2105.03023 [cs]}
  \urlprefix\url{http://arxiv.org/abs/2105.03023}.
\newblock ArXiv: 2105.03023.

\bibitem[{Madotto et~al.(2020)Madotto, Ishii, Lin, Dathathri, and
  Fung}]{madotto_plug-and-play_2020}
Madotto, A.; Ishii, E.; Lin, Z.; Dathathri, S.; and Fung, P. 2020.
\newblock Plug-and-{Play} {Conversational} {Models}.
\newblock \emph{arXiv:2010.04344 [cs]}
  \urlprefix\url{http://arxiv.org/abs/2010.04344}.
\newblock ArXiv: 2010.04344.

\bibitem[{Mai et~al.(2020)Mai, Pappas, Montero, Smith, and
  Henderson}]{mai_plug_2020}
Mai, F.; Pappas, N.; Montero, I.; Smith, N.~A.; and Henderson, J. 2020.
\newblock Plug and {Play} {Autoencoders} for {Conditional} {Text} {Generation}.
\newblock In \emph{Proceedings of the 2020 {Conference} on {Empirical}
  {Methods} in {Natural} {Language} {Processing} ({EMNLP})}, 6076--6092.
  Online: Association for Computational Linguistics.
\newblock \doi{10.18653/v1/2020.emnlp-main.491}.
\newblock \urlprefix\url{https://www.aclweb.org/anthology/2020.emnlp-main.491}.

\bibitem[{Mostafazadeh et~al.(2016)Mostafazadeh, Chambers, He, Parikh, Batra,
  Vanderwende, Kohli, and Allen}]{mostafazadeh_corpus_2016}
Mostafazadeh, N.; Chambers, N.; He, X.; Parikh, D.; Batra, D.; Vanderwende, L.;
  Kohli, P.; and Allen, J. 2016.
\newblock A {Corpus} and {Evaluation} {Framework} for {Deeper} {Understanding}
  of {Commonsense} {Stories}.
\newblock \emph{Proceedings of the 2016 Conference of the North \{A\}merican
  Chapter of the Association for Computational Linguistics: Human Language
  Technologies} 839--849.
\newblock \urlprefix\url{https://www.aclweb.org/anthology/N16-1098}.
\newblock ArXiv: 1604.01696.

\bibitem[{Pascual et~al.(2020)Pascual, Egressy, Bolli, and
  Wattenhofer}]{pascual_directed_2020}
Pascual, D.; Egressy, B.; Bolli, F.; and Wattenhofer, R. 2020.
\newblock Directed {Beam} {Search}: {Plug}-and-{Play} {Lexically} {Constrained}
  {Language} {Generation}.
\newblock \emph{arXiv:2012.15416 [cs]}
  \urlprefix\url{http://arxiv.org/abs/2012.15416}.
\newblock ArXiv: 2012.15416.

\bibitem[{Rashkin et~al.(2020)Rashkin, Celikyilmaz, Choi, and
  Gao}]{rashkin-etal-2020-plotmachines}
Rashkin, H.; Celikyilmaz, A.; Choi, Y.; and Gao, J. 2020.
\newblock {P}lot{M}achines: Outline-Conditioned Generation with Dynamic Plot
  State Tracking.
\newblock In \emph{Proceedings of the 2020 Conference on Empirical Methods in
  Natural Language Processing (EMNLP)}, 4274--4295. Online: Association for
  Computational Linguistics.
\newblock \doi{10.18653/v1/2020.emnlp-main.349}.
\newblock \urlprefix\url{https://www.aclweb.org/anthology/2020.emnlp-main.349}.

\bibitem[{Wallace et~al.(2019)Wallace, Feng, Kandpal, Gardner, and
  Singh}]{wallace_universal_2019}
Wallace, E.; Feng, S.; Kandpal, N.; Gardner, M.; and Singh, S. 2019.
\newblock Universal {Adversarial} {Triggers} for {Attacking} and {Analyzing}
  {NLP}.
\newblock \emph{Proceedings of the 2019 Conference on Empirical Methods in
  Natural Language Processing and the 9th International Joint Conference on
  Natural Language Processing (EMNLP-IJCNLP)} (2019): 2153--2162.
\newblock \urlprefix\url{http://arxiv.org/abs/1908.07125}.
\newblock ArXiv: 1908.07125.

\bibitem[{Wang, Durrett, and Erk(2020)}]{wang_narrative_2020}
Wang, S.; Durrett, G.; and Erk, K. 2020.
\newblock Narrative {Interpolation} for {Generating} and {Understanding}
  {Stories}.
\newblock \emph{arXiv:2008.07466 [cs]}
  \urlprefix\url{http://arxiv.org/abs/2008.07466}.
\newblock ArXiv: 2008.07466.

\bibitem[{Yao et~al.(2019)Yao, Peng, Weischedel, Knight, Zhao, and
  Yan}]{yao_plan-and-write_2019}
Yao, L.; Peng, N.; Weischedel, R.; Knight, K.; Zhao, D.; and Yan, R. 2019.
\newblock Plan-{And}-{Write}: {Towards} {Better} {Automatic} {Storytelling}.
\newblock \emph{Proceedings of the AAAI Conference on Artificial Intelligence}
  33(1): 7378--7385.
\newblock \urlprefix\url{http://arxiv.org/abs/1811.05701}.
\newblock ArXiv: 1811.05701.

\end{thebibliography}
\end{document}